\documentclass[sigconf]{acmart}

\usepackage{graphicx}%
\usepackage{multirow}%
\usepackage{amsmath,amsfonts}%
\usepackage{amsthm}%
\usepackage{mathrsfs} 
\usepackage[title]{appendix}%
\usepackage{xcolor}%
\usepackage{textcomp}%
\usepackage{manyfoot}%
\usepackage{booktabs}%
\usepackage{algorithm}%
\usepackage{algorithmicx} 
\usepackage{algpseudocode} 
\usepackage{listings}%
\usepackage{tikz} 
\usetikzlibrary{calc}
\usepackage{pifont}
\usepackage{tabularx} 
\usepackage{multirow}
\usepackage{fontawesome}
\usepackage{subfig}

\usepackage{array}
\usepackage{makecell}
\usepackage{enumitem}
\usepackage{balance}

\newcommand{\cmark}{\ding{51}}
\newcommand{\xmark}{\ding{55}}

\AtBeginDocument{%
  \providecommand\BibTeX{{%
    Bib\TeX}}}

\copyrightyear{2026}
\acmYear{2026}
\setcopyright{cc}
\setcctype{by}
\acmConference[MM '26]{Proceedings of the 34th ACM International Conference on Multimedia}{November 10--14, 2026}{Rio de Janeiro, Brazil}
\acmBooktitle{Proceedings of the 34th ACM International Conference on Multimedia (MM '26), November 10--14, 2026, Rio de Janeiro, Brazil}
\acmDOI{10.1145/3767308.3835251}
\acmISBN{979-8-4007-2213-4/2026/11}

\begin{document}

\title{Multi-View Structural Representations for Physically-Consistent TeX Generation}

\author{Daoheng Li}
\email{lidaoheng@njust.edu.cn}
\affiliation{%
  \institution{Nanjing University of Science and Technology}
  \city{Nanjing}
  \state{Jiangsu}
  \country{China}
}

\author{Jianjun Qian*}
\thanks{*Corresponding authors.}
\email{csjqian@njust.edu.cn}
\affiliation{%
  \institution{Nanjing University of Science and Technology}
  \city{Nanjing}
  \state{Jiangsu}
  \country{China}
}

\author{Mengkai Yan}
\email{ymk@hhu.edu.cn}
\affiliation{%
  \institution{Hohai University}
  \city{Nanjing}
  \state{Jiangsu}
  \country{China}
}

\author{Jian Yang}
\email{csjyang@njust.edu.cn}
\affiliation{%
  \institution{Nanjing University of Science and Technology}
  \city{Nanjing}
  \state{Jiangsu}
  \country{China}
}

\renewcommand{\shortauthors}{Daoheng Li, Jianjun Qian, Mengkai Yan, and Jian Yang}

\begin{abstract}

Infrared and visible images provide complementary cues for low-light image generation, and physics-based methods can better exploit this complementarity. 
However, existing physics-based methods still rely heavily on paired data.
To address these limitations, we propose Multi-View Routing Mixture-of-Experts (MVR-MoE), a framework for physically consistent TeX generation that learns multi-view structural representations.
MVR-MoE mines intrinsic structural cues from infrared and visible images to organize them into multi-view representations, while the proposed multi-view hyper connections expand the feature space and preserve both shared evidence and modality-specific details.
A hierarchical routing mechanism further regulates feature utilization at both the structural-view level and the expert level, enabling adaptive exploitation of heterogeneous cues. 
The predicted temperature, emissivity, and texture maps are constrained by the Stefan-Boltzmann law to ensure physically consistent.
Extensive experiments on challenging datasets demonstrate that MVR-MoE achieves superior robustness and discriminability while remaining lightweight.

\end{abstract}



\maketitle

\section{Introduction}

\begin{figure}[!t]
\centering
\includegraphics[width=0.86\linewidth]{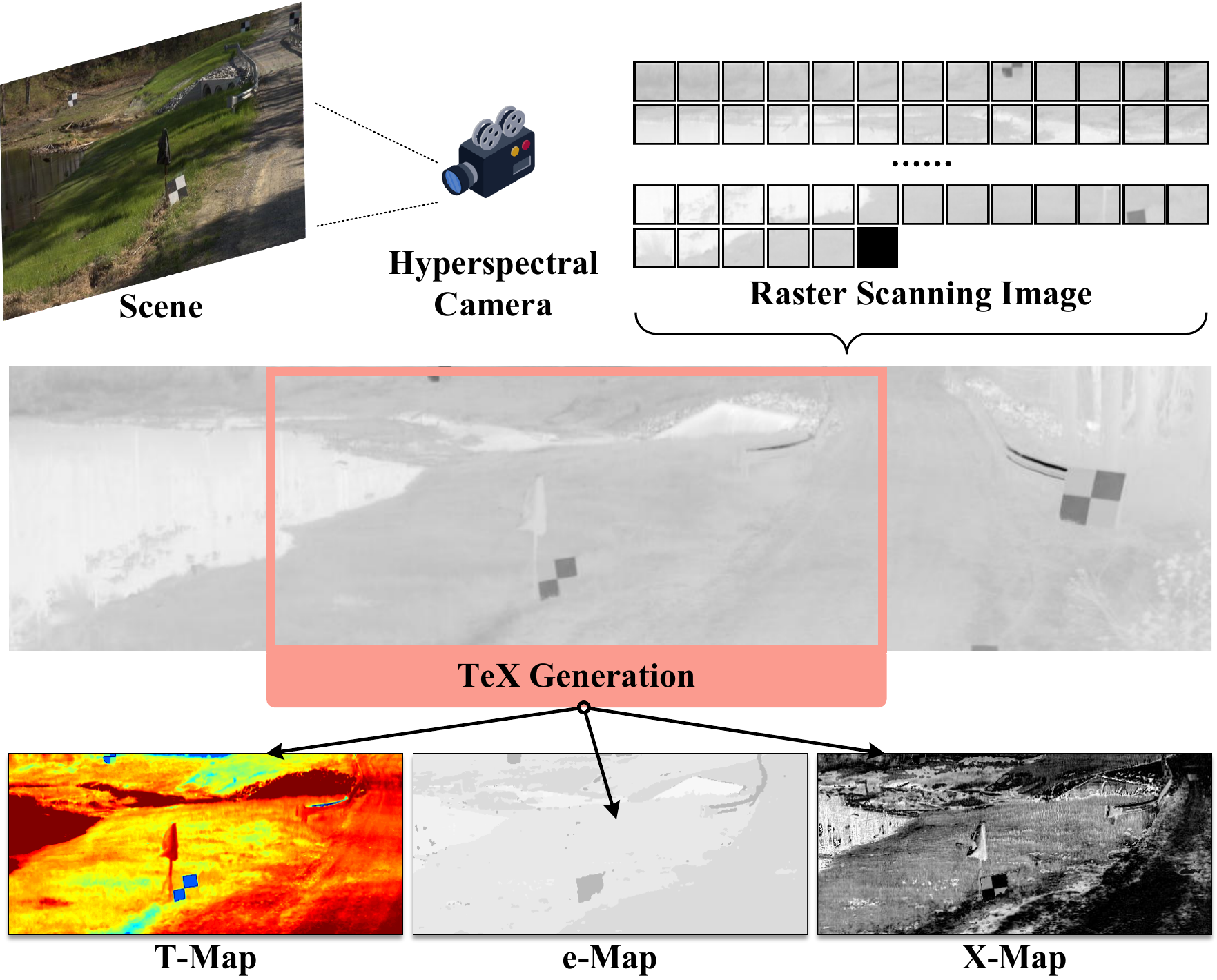}
\caption{
Illustration of infrared radiation decomposition for TeX generation.
The scene is factorized into temperature (T-Map), material emissivity (e-Map), and texture (X-Map), revealing physically meaningful distinctions that are not obvious in raw infrared measurements.
}
\label{Pic_Introduction}
\end{figure}

Low-light conditions pose significant challenges for computer vision tasks~\cite{li2020segmenting, Li2025CRTYOLO, zhang2025OmniFuse10994384}.
In such scenarios, infrared imaging can provide complementary thermal cues that remain observable under weak illumination, thereby enriching scene perception when visible information is degraded.
Motivated by this advantage, recent studies increasingly adopt multimodal perception, 
where visible and infrared images are jointly exploited to realize cross-modal fusion and improve target discriminability and overall robustness~\cite{cao2023multimodal, fu2023lraf, zeng2024mmi, lu2024gan, li2024deep, kang2025smlnet}.

Existing visible-infrared fusion methods have largely concentrated on saliency-oriented representation learning, using feature fusion, attention modulation, and cross-modal enhancement to improve target discriminability in degraded scenes~\cite{liu2024infrared_survey}.
While effective in emphasizing salient regions, these approaches mainly treat fusion as a perceptual aggregation problem.
As a result, the fused features may improve visual prominence, yet they may not explicitly preserve the physical correspondence between visible appearance and infrared radiation characteristics~\cite{lv2025slfusion, tang2023divfusion, dong2025towards}.
Maintaining physically consistent requires infrared observations to be aligned with underlying physical factors rather than being treated solely as visual features.

Recent works shift from perceptual fusion to physical decomposition, explicitly modeling the formation process of infrared radiation.
TeXNet~\cite{bao2023heat}, for instance, interprets infrared radiation through three intrinsic attributes of objects: temperature (T), material emissivity (e), and texture (X)~\cite{bao2023heat}.
As illustrated in Figure ~\ref{Pic_Introduction}, this decomposition reveals distinctions invisible in raw infrared data—for instance, grass and soil exhibit similar thermal responses but differ in emissivity and texture.
Compared with conventional fusion-based methods, such approaches aim to recover physically meaningful factors rather than directly aggregating features, thereby improving target characterization from the perspective of radiation decomposition.
To reduce acquisition complexity, more recent studies further explore temperature-emissivity decomposition using standard infrared sensors, enabling direct decomposition from infrared images with fewer imaging requirements.
However, these methods still underexploit the structural information inherent in source images, such as edges, layout regularities, and scale-dependent cues, which may help preserve physically consistent more effectively~\cite{hang2013temperature, ermida2024introducing, zhang2025deep}.

To overcome this limitation, we propose Multi-View Routing Mixture-of-Experts (MVR-MoE) for temperature-emissivity-texture (TeX) generation. 
MVR-MoE unifies structural diversity and physically consistent in multimodal features. 
We introduce multi-view hyper connections to expand the structural feature and enable cross-view interaction among structural representation, while preserving modality-specific characteristics in each branch. 
A hierarchical routing mechanism is employed within the mixture-of-experts module, where feature-level routing selects informative structural views and expert-level routing aggregates specialized predictions under the task context. 
MVR-MoE preserves both shared structural evidence and distinctive modality-specific details, yielding more discriminative and physically consistent representations for TeX generation.

Our main contributions are as follows:
\vspace{-0.5pt}
\begin{itemize}[
    leftmargin=10pt,
    topsep=0pt,
    labelsep=0.25em
]

\item We propose MVR-MoE for TeX generation, which leverages disentangled structural views to enforce physically consistent in generated TeX channels.

\item We design a multi-view hyper connections (MVHC) module to organize structural representations into adaptive multi-view branches, enabling more flexible cross-view interaction and view-aware routing.

\item Extensive experiments demonstrate that MVR-MoE achieves superior TeX generation performance, with improved robustness under thermal degeneracy and modality imbalance while maintaining a favorable efficiency--accuracy trade-off.

\end{itemize}



\section{Related Work}

\subsection{Multimodal Low-Light Visual Perception}
Infrared and visible image fusion has been extensively studied using both generative and discriminative paradigms. 
Early generative approaches model the distribution of fused images via adversarial learning, such as FusionGAN \cite{ma2019fusiongan} and GANMcC \cite{ma2020ganmcc}, while diffusion-based methods including Dif-Fusion \cite{yue2023dif} and DDFM \cite{zhao2023ddfm} improve stability and color fidelity through iterative refinement. 
Recent works further incorporate pre-trained models, where MAFS \cite{liu2022convnet} and MaeFuse \cite{song2020denoising} exploit transferable representations for joint optimization. However, these methods tend to introduce redundant infrared information or noise under overexposed conditions.
In parallel, many methods rely on carefully designed loss functions to guide fusion, such as SDNet \cite{zhang2021sdnet}, TUFusion \cite{zhao2023tufusion}, CAMF \cite{tang2023camf}, AMSFusion \cite{zhang2020vifb}, and texture and content guided models \cite{zhang2024texture}, with Xie et al. \cite{xie2025overexposed} explicitly addressing overexposure via prior modeling. 
While efficient, these approaches are sensitive to loss design and often fail to capture cross-modal semantic complementarity.
More recent studies integrate fusion with downstream tasks including TarDal \cite{liu2022target}, SegMif \cite{liu2023multi}, TDFusion \cite{bai2025task}, and DifusionSeg \cite{wang2025difusionseg}, enabling joint optimization with detection or segmentation. 
Although effective in preserving salient structures, their performance degrades under extreme illumination due to inaccurate localization of infrared cues.
Meanwhile, increasing attention has been paid to structure-aware modeling for low-light perception. 
Methods such as SLFusion \cite{lv2025slfusion}, DIVFusion \cite{tang2023divfusion}, Structure-Guided Transformer \cite{dong2025towards}, and UMCFuse \cite{li2025umcfuse} explicitly incorporate structural priors to enhance detail preservation and robustness in challenging conditions, highlighting the critical role of structural information in multimodal low-light visual perception.

\subsection{Radiation physical decomposition}
Thermal infrared (TIR) decomposition is typically formulated as a temperature--emissivity separation (TES) problem. 
The classical ASTER TES estimates emissivity via spectral contrast and derives temperature \cite{gillespie1998temperature}, but the problem is inherently ill-posed due to their strong coupling, especially with limited spectral information \cite{payan2004analysis}.
To mitigate this issue, various TES extensions introduce additional constraints, including smoothing and nonlinear formulations \cite{miao2019temperature}, subspace and spectral regularization \cite{hang2013temperature, wang2019evaluation}, and atmospheric or other physical factors \cite{kim20212, ermida2024introducing}. 
These methods generally rely on multi-band or hyperspectral data.
With hyperspectral observations, learning-based approaches further improve estimation by modeling complex mappings from radiance to physical variables \cite{hang2013temperature, xu2021ill}, and recent works integrate deep learning with TES to enhance accuracy while maintaining physically consistent \cite{zhang2025deep}.
Overall, existing methods depend either on strongly constrained TES formulations or hyperspectral and learning-based models, limiting their applicability under constrained sensing conditions.

\section{Method}

\begin{figure*}[!t]
\centering
\includegraphics[width=0.85\textwidth]{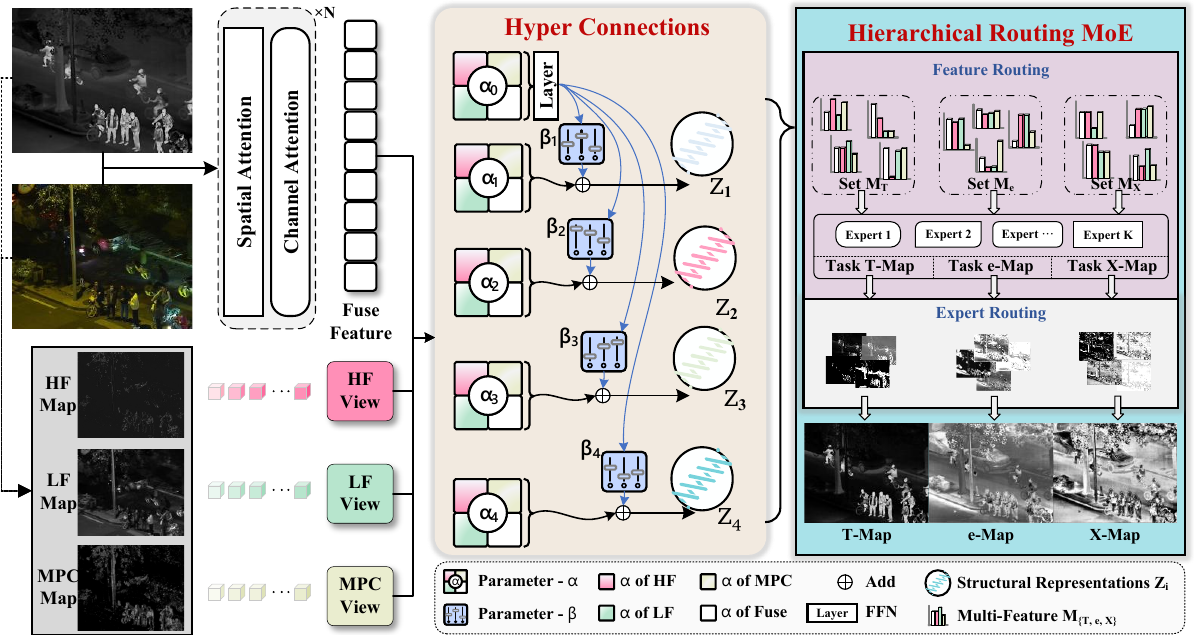}
\caption{
Overview of MVR-MoE. 
Infrared and visible images are decomposed into high-frequency (HF), low-frequency (LF), and MPC components, which jointly form a multi-view feature set.
Through multi-view hyper connections, these representations are jointly encoded into Structural Representations $Z_{1,2,3,4}$, facilitating cross-modality interactions.
Routing module performs adaptive reweighting and aggregation over $Z_{1,2,3,4}$ to construct heterogeneous representations, which are subsequently dispatched to specialized experts.
The outputs of all experts are finally integrated to generate TeX with physically consistent.
}
\label{Pic_Overview}
\end{figure*}

We propose Multi-View Routing Mixture-of-Experts (MVR-MoE), a framework for physically consistent TeX generation that disentangles shared structure and view-specific cues in multimodal representations. 
As illustrated in Figure ~\ref{Pic_Overview}, MVR-MoE takes infrared and visible images as input and jointly constructs a fused representation alongside a multi-view structural space, which captures complementary information including the high-frequency map (HF), low-frequency map (LF), and monogenic phase congruency map (MPC).
A multi-view hyper connections (MVHC) module reorganizes these structural views into context-aware branches, strengthening common patterns while preserving discriminative modality-specific evidence. 
Built upon these structural representations, the MVR-MoE employs a hierarchical routing MoE at both the feature and expert levels. 
Each generation channel integrates heterogeneous representations and leverages a shared expert pool to produce the TeX.
Robustness under thermal degeneracy and modality imbalance is improved, while over-smoothing caused by excessive fusion is mitigated and noise from unreliable cues is suppressed.
We further incorporate structural cues to support physically consistent cross-modal representation learning.

\subsection{Feature Fusion}

Given an infrared image $I_{ir}$ and a visible image $I_{vis}$, a fused feature $F$ is learned to preserve complementary characteristics of both modalities while facilitating effective cross-modal interaction. 
The module progressively models inter-modal dependencies from both spatial and channel perspectives, enabling fine-grained structural alignment as well as high-level consistency across modalities.

\noindent
\textbf{\textit{Spatial Attention}}.
Spatial interaction is performed in the frequency domain to extract local cross-modal correspondences. 
Infrared and visible features are transformed by FFT~\cite{zheng2024probing}, and cross-modal attention is computed from their frequency representations to form an initial fused feature. 
The fusion process is refined iteratively through convolutional extraction and attention-weighted interactions between the fused feature and the infrared feature. 
A residual connection with the visible feature preserves visible structural details and stabilizes feature propagation. 

\noindent
\textbf{\textit{Channel Attention}}.
Channel interaction focuses on global semantic dependencies across feature channels. 
Attention derived from the fused feature reweights the visible feature, amplifying informative channels and suppressing redundant information. 
Iterative convolution and attention modulation further strengthen semantic integration across modalities. 
A residual connection with the original $I_{vis}$ maintains modality-specific information and supports stable feature learning.

\noindent
\textbf{\textit{Fused Feature}}.
The fuse feature $F$ is dominated by infrared information and complemented by auxiliary cues from the visible modality. 
Infrared image provide the primary structural and saliency information, ensuring robust characterization of thermal targets. 
In contrast, the visible modality contributes additional texture details that are less distinguishable in the infrared domain. 
As a result, the $F$ preserves the integrity of infrared features while incorporating fine-grained appearance cues, forming a representation that balances structural reliability with detail enhancement.

\subsection{Multi-View Hyper Connections}

MVHC integrates the fused feature $F$ with disentangled structural priors.  
To bridge the modality gap without sacrificing physically consistent, we decompose the input into three complementary views: High-Frequency (HF), Low-Frequency (LF), and Monogenic Phase Congruency (MPC). 
Multiple views are formed from the same cross-modal representation under different structural constraints, enabling the network to examine scene structure from complementary viewpoints. 
The HF view captures fine-grained details, the LF view reflects coarse structures, and the MPC view encodes illumination-robust saliency.
Shared structural evidence across modalities is reinforced across these views, while modality-specific cues remain salient in individual branches. 
The resulting multi-view representation preserves both common patterns and complementary details, providing a richer basis for subsequent routing and prediction.

\noindent
\textbf{\textit{Multi-View Structural}}.
The $I_{vis}$ is decomposed into two complementary components via a transmission estimation mechanism:
\begin{equation}
L_1 = t \cdot \text{gray}(I_{vis}), \qquad
L_2 = (1-t) \cdot \text{gray}(I_{vis}),
\end{equation}
where $t$ is the estimated transmission map.
We apply guided filtering~\cite{he2012guided} to extract low-frequency structures. 
Given a guidance image $G$ and an input image $p$, the guided filter $\operatorname{GF}(G,p;r,\lambda)$ computes a filtered image $q$ by assuming a local linear model within each window $\omega_k$:
\begin{equation}
q_i = a_k G_i + b_k, \qquad i \in \omega_k,
\end{equation}
where
\begin{equation}
a_k = \frac{\operatorname{cov}_{\omega_k}(G,p)}{\operatorname{var}_{\omega_k}(G)+\lambda}, \qquad
b_k = \mu_p - a_k \mu_G.
\end{equation}
The filtered image is obtained by averaging the local coefficients over overlapping windows:
$q = \bar{a} \odot G + \bar{b}$.
In this work, we focus on the self-guided setting, where the guided filter is applied with identical guidance and input:
\begin{equation}
E_1 \!= \operatorname{GF}(L_1, L_1; r, \lambda), 
E_2 \!= \operatorname{GF}(L_2, L_2; r, \lambda), 
E_3 \!= \operatorname{GF}(I_{ir}, I_{ir}; r, \lambda).
\end{equation}
The corresponding high-frequency details are obtained by residual decomposition:
\begin{equation}
D_1 = L_1 - E_1, \qquad
D_2 = L_2 - E_2, \qquad
D_3 = I_{ir} - E_3,
\end{equation}
where $D_i$ captures edges, textures, and fine-grained structural details.
Monogenic Phase Congruency (MPC) is employed to measure structural saliency independent of illumination intensity, estimating the consistency of multi-scale filter responses.
For each detail map $D_i$, the monogenic signal is computed to obtain its local structural components. 
Specifically, let $f_{1,i}$ and $f_{2,i}$ denote the corresponding filter responses derived from the monogenic decomposition.
Based on these, the phase-congruency map is defined as
\begin{equation}
MPC_i = \frac{W_i \cdot f_{1,i} \cdot f_{2,i}}{\sum_k |R_{i,k}| + \epsilon},
\end{equation}
where $R_{i,k}$ denotes the response at scale $k$, and $\epsilon$ is a small constant for numerical stability. 
Using MPC as a structural reliability measure, the high-frequency prior is fused as
\begin{equation}
HF = \sum_{i=1}^{3} w_i D_i,
\end{equation}
with adaptive weights defined by
\begin{equation}
w_i(x) =
\frac{MPC_i(x) + \epsilon}
{\max_x \left( \sum_{j=1}^{3} MPC_j(x) \right) + \epsilon}.
\end{equation}
For low-frequency extraction, we construct an intermediate map by selecting the component with stronger low-frequency response between the visible and infrared branches:
\begin{equation}
E_4 = \mathbb{I}(E_2 > E_3)\odot E_2 + \mathbb{I}(E_2 \leq E_3)\odot E_3,
\end{equation}
where $\mathbb{I}(\cdot)$ denotes the indicator function that outputs '1' when the condition is satisfied and '0' otherwise, and $\odot$ denotes element-wise multiplication.
The final low-frequency prior is then obtained as
\begin{equation}
LF = W_4 E_4 + W_3 E_1,
\end{equation}
where $W_3$ and $W_4$ are adaptive weights determined by the structural saliency of the corresponding low-frequency maps. 
Finally, the extracted priors $\{HF, LF, MPC\}$ provide complementary structural information, including fine details, global intensity structure, and illumination-invariant geometric features.

\noindent
\textbf{\textit{Structural Representations}}.  
Hyper Connections (HC~\cite{zhu2024hyper}) mechanism is introduced to construct multiple structural views, replacing direct concatenation. 
The proposed MVHC injects image-intrinsic structural cues to produce heterogeneous representations.
The MVHC module takes a set of input features $\mathcal{S} = \{S_1, S_2, S_3, S_4\} = \{F, HF, LF, MPC\}$.
A set of learnable coefficients \(\alpha_{i,j}\) parameterizes the contribution of the \(i\)-th input feature to the \(j\)-th branch. 
Each branch feature is obtained by a weighted combination of all inputs:
\begin{equation}
\hat{F}_j = \sum_{i=1}^{4} \alpha_{i,j} S_i, 
\qquad j \in \{0,1,2,3,4\}.
\end{equation}
The base feature is generated by applying a residual transformation followed by Layer(FFN): $F_0 = \text{FFN}(\hat{F}_0)$.
Another set of learnable coefficients \(\beta_j\) modulates the interaction between the base feature $F_0$ and each branch:
\begin{equation}
Z_j = \beta_j F_0 + \hat{F}_j, 
\qquad j \in \{1,2,3,4\}.
\end{equation}
The MVHC module dynamically determines how multi-view structural representations $\mathcal{Z}={Z_1,Z_2,\dots,Z_K}$ are fused and propagated across feature branches.
Each branch emphasizes complementary structural cues derived from the input modalities and maintains the shared cross-modal feature. 
The branch-wise fusion weights are learnable, allowing common structures to be strengthened while preserving view-specific information for subsequent prediction.

\begin{figure}[!t]
\centering
\includegraphics[width=0.85\linewidth]{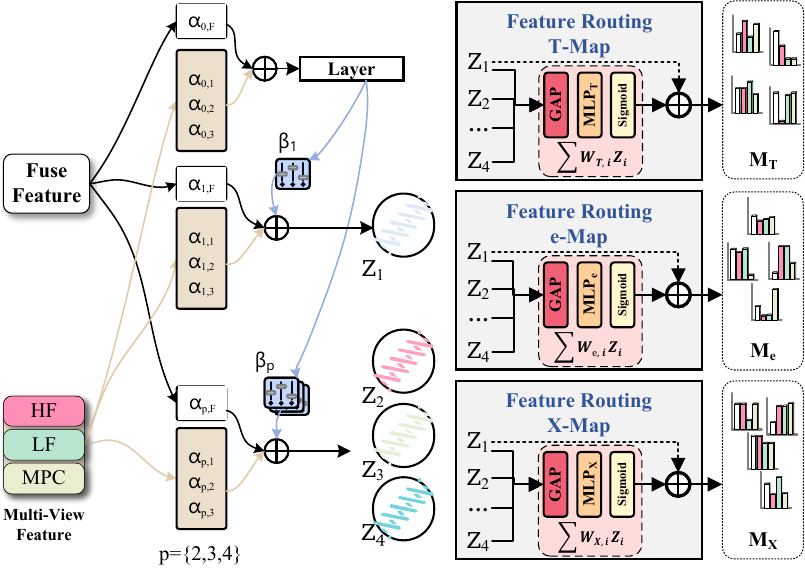}
\caption{
MVHC. 
Through hyper connections, HF, LF, MPC interact with the $F$ to form four multi-view branches that preserve both shared consensus and view-specific diversity. 
For each prediction channel, a feature-level routing module first constructs heterogeneous feature representations via adaptive composition of Structural Representations.
}
\label{Pic_MVHC}
\end{figure}

\subsection{TeX Generation}

A hierarchical routing Mixture-of-Experts is employed for TeX generation. 
We incorporate structural information into routing to better preserve physically consistent.
As shown in Figure \ref{Pic_MVHC}, the feature routing constructs heterogeneous feature for experts, assigning mixed structural features to different experts. 
The resulting expert outputs are then aggregated by gating networks to produce candidate predictions for each task channel.

\noindent
\textbf{\textit{Feature-level Routing}.}
Given the Structural Representations set ${Z_1, Z_2, Z_3, Z_4}$, we instantiate three independent feature routing, namely $\mathrm{FR}_T$, $\mathrm{FR}_e$, and $\mathrm{FR}_X$, each dedicated to a specific generation channel $c\!\in\!\{T\!,\!e\!,\!X\}$.
For each feature routing, the views ${Z_1, Z_2, Z_3, Z_4}$ are incorporated through structured composition to construct enriched and heterogeneous feature representations.
Concretely, for each auxiliary view $Z_i$, we first apply global average pooling followed by a lightweight MLP to produce a scalar routing score:
\begin{equation}
r_{c,i} = \mathrm{MLP}_c(\mathrm{GAP}(Z_i)),
\end{equation}
where $\mathrm{GAP}(\cdot)$ denotes global average pooling and $\mathrm{MLP}_c$ represents the learnable parameters of the feature routing for channel $c$.
Each score is independently transformed into a gating weight via a Sigmoid function $w_{c,i} = \sigma(r_{c,i})$, and the heterogeneous feature for channel $c$ is computed as
\begin{equation}
M_c = \sum_{i=1}^{4} w_{c,i} Z_i.
\end{equation}
Such a routing strategy allows each generation channel to adaptively constructs heterogeneous features while preserving a stable base representation, thereby improving feature specialization and reducing unnecessary feature interference.

The heterogeneous features are then processed by a shared expert pool consisting of K experts $\mathcal{E} = \{E_1, E_2, \dots, E_K\}$.
Each expert independently transforms the heterogeneous feature to produce candidate outputs:
\begin{equation}
Map_c^{(k)} = E_k(M_c),
\end{equation}
where $Map_c^{(k)}$ denotes the output generated by expert k for channel c.
The expert pool is shared across all channels, allowing experts to specialize in different structural patterns while supporting multiple generation tasks.

\noindent
\textbf{\textit{Expert-level Routing}}.
To route channel-specific information to the corresponding prediction branch, we introduce channel-specific gating networks $\{\text{Gate}_T, \text{Gate}_e, \text{Gate}_X\}$.
For each channel $c$, the gating network generates routing scores from the pooled channel representation:
\begin{equation}
\mathbf{s}_c =
U_c \cdot \text{Pool}(M_c),
\end{equation}
where $U_c$ denotes the parameters of the gating network.
The routing weights are then obtained by applying a Top-$K$ softmax operator:
\begin{equation}
\mathbf{g}_c = \text{TopK}\big(\text{Softmax}(\mathbf{s}_c)\big),
\end{equation}
which retains the $K$ highest-weight experts and normalizes their contributions.
The routed expert features are aggregated as
\begin{equation}
\mathbf{F}_c =
\sum_{k \in \mathcal{S}_c}
g_{c,k}\mathbf{F}_c^{(k)},
\end{equation}
where $\mathcal{S}_c$ denotes the Top-$K$ expert set and $\mathbf{F}_c^{(k)}$ is the feature output of the $k$-th expert.
Finally, a channel-specific projection head maps the aggregated feature to the single-channel prediction:
\begin{equation}
Map_c = \rho_c(\mathbf{F}_c).
\end{equation}
TeX generation can be summarized as a nested routing operation
\begin{equation}
Map_c =
\sum_{k=1}^{K}
g_{c,k}
E_k
\left(
\sum_{i=1}^{4}
s_{c,i} Z_i
\right),
\end{equation}
where the feature routing determines how structural views are composed, while the expert routing determines which expert outputs dominate the final prediction.
The proposed architecture employs hierarchical routing to decouple structural view composition, feature routing organizes multi-view features, and GateNet modules weight expert outputs. 
Cross-modal consensus and view-specific cues are jointly retained, supporting adaptive expert specialization for robust TeX generation.

\subsection{Loss Function}

We introduce physics-guided constraints during training to encourage physically consistent generation of the channels rather than superficial visual plausibility. 
Specifically, the loss design incorporates the Stefan–Boltzmann radiation model together with channel consistency constraints, enabling the generated T, e, and X maps to satisfy physically relationships.

\textbf{\textit{Stefan–Boltzmann Constraint}}.
Based on Stefan–Boltzmann law, we impose a reconstruction constraint that couples the generated channels T,e,X with the observed infrared image.
The reconstructed infrared intensity is formulated as
\begin{equation}
\hat{I}_{ir}(x,y)
=
e(x,y)\,\sigma\,T(x,y)^4
+
\bigl(1-e(x,y)\bigr)X(x,y),
\end{equation}
where T(x,y) denotes the temperature map, e(x,y) denotes the emissivity map, X(x,y) represents the texture, and $\sigma$ is the Stefan–Boltzmann constant.
The reconstruction loss is defined to match the reconstructed infrared intensity $\!\hat{I}_{ir}\!$ with the observed infrared image $\!I_{ir}$: 
\begin{equation}
\mathcal{L}_{\text{re}}
=
\left\|
\hat{I}_{ir} - I_{ir}
\right\|_1
+
D_{KL}
\bigl(
P(I_{ir})
\parallel
P(\hat{I}_{ir})
\bigr),
\end{equation}
the first term preserves pixel-level fidelity, whereas the second term encourages statistical consistency between the reconstructed and observed infrared intensities.

\textbf{\textit{Channel Consistency Loss}}.
The temperature T, emissivity e, and texture X represent different physical aspects of the same object. 
These quantities are governed by the intrinsic properties of materials.
Accordingly, although their values differ across channels, their spatial structures remain consistent.
In particular, when materials with different emissivities form a boundary in the image plane, the boundary is often accompanied by variations in emissivity, temperature, and texture.
Therefore, the spatial structures of T, e, and X exhibit strong structural alignment.
The channel consistency loss is defined in terms of high-frequency and low-frequency components:
\begin{equation}
\begin{aligned}
\mathcal{L}_{\text{e-T/X}}
=
\frac{1}{H W}
& 
\left( 
\left\| T_{\text{HF}} - e_{\text{HF}} \right\|_1
+
\left\| X_{\text{HF}} - e_{\text{HF}} \right\|_1
\right)
\\
+
&\left\| T_{\text{LF}} - e_{\text{LF}} \right\|_1
+
\left\| X_{\text{LF}} - e_{\text{LF}} \right\|_1 
,
\end{aligned}
\end{equation}
where $H$ and $W$ denote the spatial dimensions of the feature maps. 
The subscripts HF and LF denote the high-frequency and low-frequency components extracted from the respective channels.

\textbf{\textit{Temperature–Texture Complementarity}}.
In multimodal imagery, high-intensity regions may arise from either elevated temperature or strong illumination and reflectance effects. 
A complementary relationship between the temperature channel T and the texture channel X is therefore imposed during generation.
The constraint is built on the normalized infrared intensity $I^{norm}_{IR}$ and the blurred visible luminance map $Y_{blur}$. 
Their difference provides a heuristic indication of whether a bright region is more likely attributable to thermal radiation or illumination.
The complementarity loss is defined as
\begin{equation}
\mathcal{L}_{\text{T-X}}
=
\left\|
T -
\phi
\left(
I_{IR}^{norm} - Y_{blur}
\right)
\right\|_2^2
+
\left\|
X -
\phi
\left(
Y_{blur} - I_{IR}^{norm}
\right)
\right\|_2^2 ,
\end{equation}
where $\phi(\cdot)$ denotes the sigmoid function.
This constraint encourages the model to assign bright regions either to temperature or to illumination effects, thereby reducing ambiguity between the T and X channels.

\textbf{\textit{MoE Gate Task Loss.}}
To improve expert utilization in the multi-task MoE module, we introduce a gate-level regularization term that encourages balanced expert assignment within each task. 
Let $\mathcal{T}\!=\!\{T\!,\!e\!,\!X\}$ denote the task set, $B$ the batch size, and $K$ the number of experts. For each task $t \!\in\! \mathcal{T}$, the gating output is $\mathbf{G}^{(t)}\! \in\! \mathbb{R}^{B \times K}$, where $\mathbf{G}^{(t)}_{b,i}$ denotes the weight assigned to expert $i$ for sample $b$.
For each expert, we compute the coefficient of variation over the batch dimension:
\begin{equation}
\mathrm{CV}^{(t)}_i = \frac{\sigma^{(t)}_i}{\mu^{(t)}_i + \epsilon},
\end{equation}
where
\begin{equation}
\mu^{(t)}_i = \frac{1}{B}\sum_{b=1}^{B}\mathbf{G}^{(t)}_{b,i}, 
\qquad
\sigma^{(t)}_i = \sqrt{\frac{1}{B}\sum_{b=1}^{B}\left(\mathbf{G}^{(t)}_{b,i}-\mu^{(t)}_i\right)^2}.
\end{equation}
The load balancing loss is defined as
\begin{equation}
\mathcal{L}_{\mathrm{gate}}
=
\frac{1}{|\mathcal{T}|}
\sum_{t \in \mathcal{T}}
\frac{1}{K}
\sum_{i=1}^{K}
\left(\mathrm{CV}^{(t)}_i\right)^2.
\end{equation}
Furthermore, we constrain the extent to which an expert is shared across tasks. 
We define a binary usage indicator:
\begin{equation}
U^{(t)}_{b,i} =
\begin{cases}
1, & \mathbf{G}^{(t)}_{b,i} > 0, \\
0, & \text{otherwise},
\end{cases}
\end{equation}
and compute the task usage count of expert $i$ as
\begin{equation}
C_i = \frac{1}{B}\sum_{b=1}^{B}\left(U^{(T)}_{b,i}+U^{(e)}_{b,i}+U^{(X)}_{b,i}\right).
\end{equation}
We penalize experts that are selected by more than two tasks:
\begin{equation}
\mathcal{L}_{\mathrm{pertask}}
=
\frac{1}{K}\sum_{i=1}^{K}\max(0, C_i - 2).
\end{equation}

\textbf{\textit{Overall Physically Consistent Objective}}.
Combining the above components, the final loss function is defined as
\begin{equation}
\begin{aligned}
\mathcal{L}
=
\mathcal{L}_{\text{re}}
+
\lambda_1 \mathcal{L}_{\text{e-T/X}}
+
\lambda_2 \mathcal{L}_{\text{T-X}}
+
\lambda_{\mathrm{3}}\mathcal{L}_{\mathrm{gate}}
+
\lambda_{\mathrm{4}}\mathcal{L}_{\mathrm{pertask}},
\end{aligned}
\end{equation}
where $\lambda_1$, $\lambda_2$, $\lambda_3$ and $\lambda_4$ are weighting coefficients.
The proposed objective enforces physically consistent across the generated channels, ensuring that the predicted temperature, emissivity, and texture maps remain consistent with both the observed infrared signal and the intrinsic physical relationships among these quantities.

\section{Experiment}

\begin{table}[!t]
\centering
\caption{Object detection performance comparison of MVR-MoE and competing methods on the LLVIP dataset.}
\label{Table_LLVIP}
\renewcommand{\arraystretch}{0.88}
\begin{tabular*}{\linewidth}{@{\extracolsep{\fill}}lcccc}
    \hline
    \textbf{Method}
    & \textbf{Modality}
    & $mAP_{50}$
    & \multicolumn{1}{c}{$mAP_{75}$}
    & \multicolumn{1}{c}{$mAP_{50:95}$}
    \\ \hline
    YoLov10 \cite{wang2024yolov10}
    & RGB 
    & 90.0  & 57.3  & 52.7
    \\ 
    YoLov10 \cite{wang2024yolov10}
    & IR 
    & 96.1  & 74.8  & 64.7
    \\ \hline
    MMI-Det \cite{zeng2024mmi}
    & IR+RGB
    & 98.9  & 73.5  & 64.4
    \\
    GAN-HA \cite{lu2024gan}
    & IR+RGB
    & 94.3  & N/A   & 56.0
    \\
    RSDet \cite{zhao2024removal}
    & IR+RGB
    & 95.8  & 70.4  & 61.3
    \\
    ICAFusion \cite{shen2024icafusion}
    & IR+RGB
    & N/A   & N/A   & N/A 
    \\
    UniRGB-IR \cite{yuan2024unirgb}
    & IR+RGB
    & 96.1  & 72.2  & 63.2
    \\
    Fusion-Mamba \cite{dong2024fusion}
    & IR+RGB
    & 97.0  & N/A   & 64.3
    \\
    Diff-IF \cite{yi2024diff}
    & IR+RGB
    & 93.3	& 68.4	& 59.5
    \\
    DDFM \cite{zhao2023ddfm}
    & IR+RGB
    & 92.1	& 65.5	& 59.2
    \\
    Text-IF \cite{yi2024text}
    & IR+RGB
    & 94.1	& 67.6	& 60.2
    \\
    FusionBooster \cite{cheng2025fusionbooster}
    & IR+RGB 
    & 97.4  & N/A   & 69.3  
    \\
    FD$^{2}$-Net \cite{li2025fd2} 
    & IR+RGB 
    & 96.2  & 70.0  & N/A 
    \\
    M$^{2}$D-LIF \cite{zhao2025rethinking} 
    & IR+RGB 
    & N/A   & N/A   & 70.8 
    \\
    IC-Fusion \cite{hwang2025multispectral}
    & IR+RGB 
    & 97.9  & 81.5  & 70.3
    \\
    RCS-MD \cite{shao2025representation} 
    & IR+RGB 
    & 97.7  & 81.7  & 69.5
    \\
    EI$^2$Det \cite{hu2025ei} 
    & IR+RGB 
    & 98.0  & 73.2  & 63.9
    \\
    MCOR \cite{jang2025multispectral}
    & IR+RGB 
    & 97.6  & 73.7  & 64.9
    \\
    LUT-Fuse \cite{yi2025lut}
    & IR+RGB 
    & 94.1  & N/A   & 61.4
    \\
    WaveMamba \cite{zhu2025wavemamba}
    & IR+RGB
    & \underline{\textbf{98.3}}  & N/A   & 66.0
    \\
    OmniFuse \cite{zhang2025omnifuse}
    & IR+RGB
    & 92.1  & 74.6  & 63.5
    \\
    TDHF \cite{li2025tdhf}
    & IR+RGB
    & 97.8  & 82.1  & 69.9
    \\ \hline
    \textbf{MVR-MoE}
    & IR+RGB
    & 98.2  & \underline{\textbf{84.0}}  & \underline{\textbf{72.0}} 
    \\ \hline
\end{tabular*}
\end{table}

\begin{table}[!t]
\centering
\caption{Object detection performance comparison of MVR-MoE and competing methods on the FLIR dataset.}
\label{Table_FLIR}
\renewcommand{\arraystretch}{0.88}
\begin{tabular*}{\linewidth}{@{\extracolsep{\fill}}lcccc}
    \hline
    \textbf{Method}
    & \textbf{Modality}
    & $mAP_{50}$
    & \multicolumn{1}{c}{$mAP_{75}$}
    & \multicolumn{1}{c}{$mAP_{50:95}$}
    \\ \hline
    YoLov10 \cite{wang2024yolov10}
    & RGB 
    & 66.3  & 27.7  & 32.7
    \\ 
    YoLov10 \cite{wang2024yolov10}
    & IR 
    & 78.2  & 39.6  & 42.3
    \\ \hline
    MMI-Det \cite{zeng2024mmi}
    & IR+RGB
    & 79.8  & 35.8  & 40.5
    \\
    RSDet \cite{zhao2024removal}
    & IR+RGB
    & 83.9  & 40.1  & 43.8 
    \\
    ICAFusion \cite{shen2024icafusion}
    & IR+RGB
    & 79.2  & 36.9  & 41.4
    \\
    UniRGB-IR \cite{yuan2024unirgb}
    & IR+RGB
    & 81.4  & 40.2  & 44.1 
    \\
    Fusion-Mamba \cite{dong2024fusion}
    & IR+RGB
    & 84.9  & 45.9   & 47.0 
    \\
    MMFN \cite{yang2024multidimensional}
    & IR+RGB 
    & 80.8  & 36.6  & 41.7 
    \\
    FD$^{2}$-Net \cite{li2025fd2} 
    & IR+RGB 
    & 82.9  & 42.5  & N/A 
    \\
    M$^{2}$D-LIF \cite{zhao2025rethinking} 
    & IR+RGB 
    & N/A   & N/A   & 46.1 
    \\
    IC-Fusion \cite{hwang2025multispectral}
    & IR+RGB 
    & 83.3  & 42.6  & 46.1    
    \\
    RCS-MD \cite{shao2025representation} 
    & IR+RGB 
    & 81.5  & 47.2  & 47.8   
    \\
    EI$^2$Det \cite{hu2025ei} 
    & IR+RGB 
    & 80.2  & N/A   & N/A   
    \\
    MCOR \cite{jang2025multispectral}
    & IR+RGB 
    & 78.2  & 34.4  & 39.9  
    \\
    TDHF \cite{li2025tdhf}
    & IR+RGB
    & 84.9  & 46.7  & 48.2 
    \\ \hline
    \textbf{MVR-MoE}
    & IR+RGB
    & \underline{\textbf{85.7}}	& \underline{\textbf{53.2}}	& \underline{\textbf{51.8}} 
    \\ \hline
\end{tabular*}
\end{table}

\begin{table*}[!t]
\centering
\caption{Quantitative comparison of MVR-MoE and other methods on the M3FD dataset for object detection, including per-class detection performance.}
\label{Table_M3FD}
\renewcommand{\arraystretch}{0.88}
\begin{tabular*}{\linewidth}{@{\extracolsep{\fill}}lccccccccc}
\hline
    Method  & Modality  & People    & Car    & Bus    & Motorcycle     & Lamp    & Truck
    & $mAP_{50}$    & $mAP_{50:95}$
    \\ \hline
    YoLov10 \cite{wang2024yolov10}
    & RGB   
    & 44.5   & 68.7   & 75.4   & 49.3   & 50.1   & 65.8   
    & 86.1   & 59.0  
    \\ 
    YoLov10 \cite{wang2024yolov10}
    & IR
    & 53.1   & 64.6   & 72.4   & 41.1   & 47.7   & 61.4   
    & 84.5   & 56.7  
    \\ \hline
    CFT \cite{qingyun2022cross}
    & IR+RGB
    & 52.0   & 68.2   & 79.2   & 49.9   & 45.2   & \textbf{69.6}     
    & 89.8   & 60.7 
    \\ 
    TarDAL \cite{liu2022target}
    & IR+RGB
    & 49.8   & 65.4   & 69.5   & 46.6   & 43.7   & 61.1   
    & 86.0   & 56.0 
    \\ 
    MetaFusion \cite{zhao2023metafusion}
    & IR+RGB
    & 48.4   & 66.7   & 70.5   & 49.1   & 46.4   & 59.9     
    & 86.7   & 56.8 
    \\ 
    CDDFuse \cite{zhao2023cddfuse}
    & IR+RGB
    & 46.1   & 69.7   & 74.2   & 42.2   & 44.2   & 62.7   
    & 87.0   & 56.5 
    \\ 
    ICAFusion \cite{shen2024icafusion}
    & IR+RGB
    & 48.8   & 68.5   & 72.3   & 45.5   & 43.6   & 64.7   
    & 87.4   & 57.2 
    \\
    Text-IF \cite{yi2024text}
    & IR+RGB
    & 49.1   & 69.0   & 40.3   & 27.9   & 24.1   & 32.1   
    & N/A    & 40.4 
    \\
    MTG-Fusion \cite{wang2025multi}
    & IR+RGB
    & \textbf{59.5}   &\textbf{ 75.2}   & 50.5   & 41.0   & 34.8   & 54.8   
    & N/A    & 52.6 
    \\
    CoDAF \cite{zongzhen2025cross}
    & IR+RGB
    & 46.8   & 69.4   & 78.0   & \textbf{54.2}   & 50.7   & 68.4   
    & \textbf{90.8}   & 61.2
    \\ \hline
    \textbf{MVR-MoE}
    & IR+RGB
    & \textbf{59.5}   & 70.1   & \textbf{80.3}   & 46.2   & \textbf{55.5}   & 68.9
    & 88.8   & \textbf{63.4} 
    \\ \hline
\end{tabular*}
\end{table*}

\subsection{Implementation Setting}
\textbf{\textit{Datasets}}.
We evaluate our method on low-light object detection datasets: FLIR\cite{flir2021}, LLVIP\cite{jia2021llvip}, and M3FD\cite{liu2022target}. 
These datasets provide aligned IR-VIS image pairs and cover diverse real-world scenes. 

\noindent
\textbf{\textit{Metric}}.
We use mean Average Precision (mAP) as the evaluation metric, including mAP@0.5, mAP@0.75, and mAP@0.5:0.95 with a 0.05 step size. 
The models for each downstream task were retrained on the TeX.
We compare computational efficiency using three metrics: average runtime, FLOPs, and the number of parameters. 
Average runtime measures the time required to complete a task and reflects inference speed. 
FLOPs quantify the computational cost during forward propagation. 
The number of parameters indicates the model size and complexity, which are closely related to memory and storage overhead.

\subsection{Comparison with State-of-the-art Methods}
The comparison results are reported in Tables~\ref{Table_LLVIP}, \ref{Table_FLIR}, and \ref{Table_M3FD}.
The compared methods include both image-fusion-based pipelines and end-to-end multispectral detection approaches.

\noindent
\textbf{Results on LLVIP.}
As shown in Table~\ref{Table_LLVIP}, the proposed MVR-MoE achieves the best performance across all evaluation metrics, reaching 98.2\% $mAP_{50}$, 84.0\% $mAP_{75}$, and 72.0\% $mAP_{50:95}$.
Compared with the strongest existing methods such as IC-Fusion and TDHF, our approach provides consistent improvements, especially under the stricter metrics $mAP_{75}$ and $mAP_{50:95}$. 
These results demonstrate that the proposed framework effectively exploits complementary multispectral information and improves localization accuracy under challenging illumination conditions.

\noindent
\textbf{Results on FLIR.}
Table~\ref{Table_FLIR} reports the detection performance on the FLIR dataset.
MVR-MoE again achieves the best results across all three metrics, obtaining 85.7\% $mAP_{50}$, 53.2\% $mAP_{75}$, and 51.8\% $mAP_{50:95}$.
Compared with recent state-of-the-art methods such as TDHF and Fusion-Mamba, our approach shows notable improvements, particularly on the more challenging metrics. 
These results indicate that the proposed architecture maintains strong robustness and effectiveness across different multispectral datasets.

\noindent
\textbf{Results on M3FD.}
The comparison results on the M3FD dataset are presented in Table~\ref{Table_M3FD}. 
MVR-MoE achieves competitive overall performance with 88.8\% $mAP_{50}$ and 63.4\% $mAP_{50:95}$, outperforming most existing fusion-based methods while remaining comparable to recent end-to-end approaches. 
In addition, the proposed method shows stable detection performance across multiple object categories, particularly for challenging classes such as lamps and pedestrians. 
These results suggest that the multi-view feature evolution and expert routing mechanisms help the model capture complementary multispectral cues for diverse object categories.

To evaluate the generalizability of MVR-MoE, we conduct cross-dataset transfer experiments by training the model on one dataset and directly generating TeX representations on another dataset without target-domain fine-tuning. 
Across FLIR, LLVIP, and M3FD, MVR-MoE consistently achieves stable downstream detection performance, with mAP50:95 ranging from 51.4\% to 51.6\%, 70.7\% to 72.0\%, and 62.7\% to 63.4\%, respectively. 
The small performance variation across different training-testing combinations indicates that MVR-MoE learns transferable thermal structural priors that are robust to dataset shifts.
Overall, the experimental results across three datasets demonstrate that the proposed MVR-MoE framework consistently improves multispectral detection performance and generalizes well to different scenarios.

\begin{figure*}[!t]
\centering
\setlength{\tabcolsep}{2pt}
\renewcommand{\arraystretch}{1}
\begin{tabular}{@{}cccccc@{}}
\begin{tabular}{c}
\subfloat[Visible image]{
\includegraphics[width=0.14\linewidth]{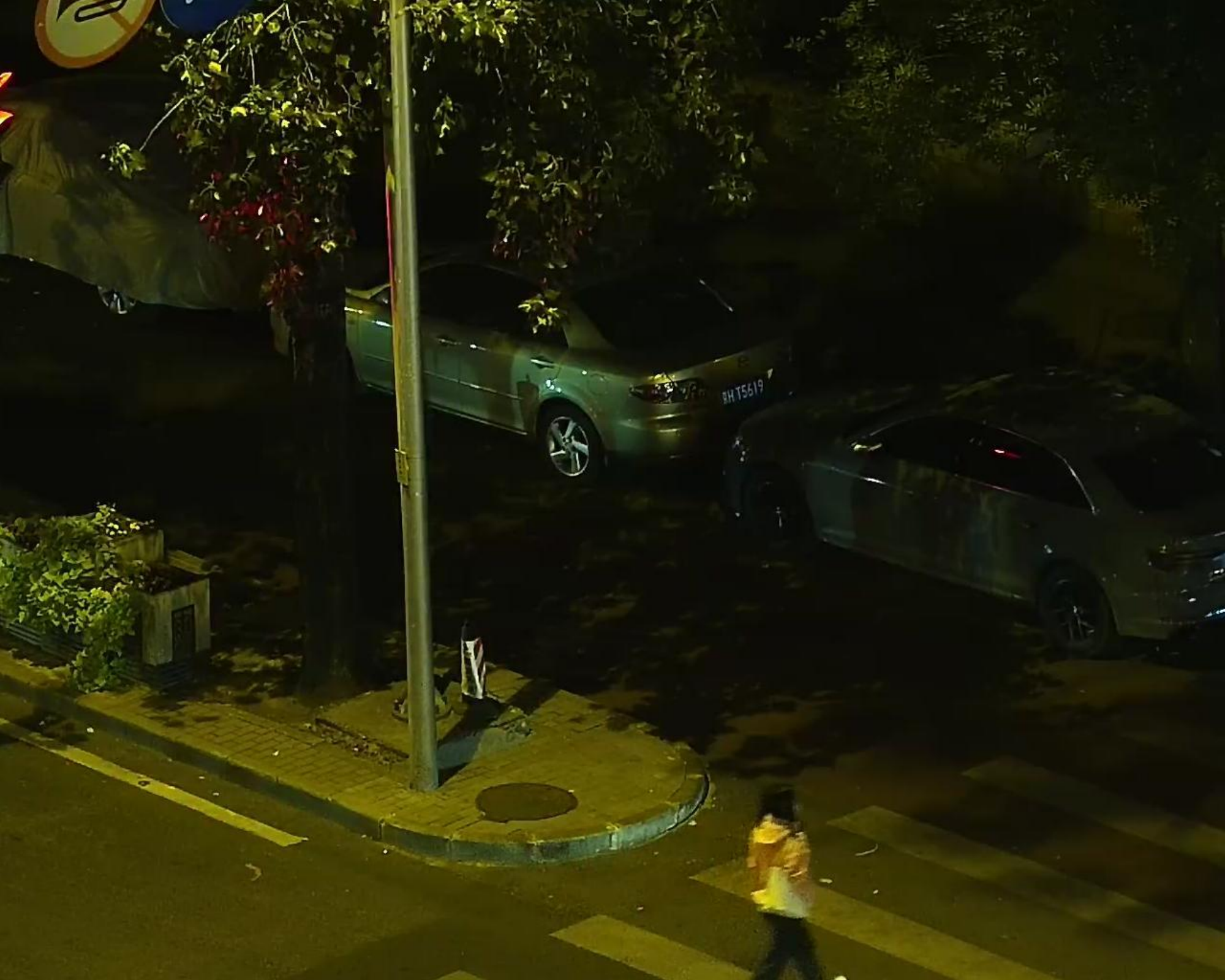}}\\[2mm]
\subfloat[Infrared image]{
\includegraphics[width=0.14\linewidth]{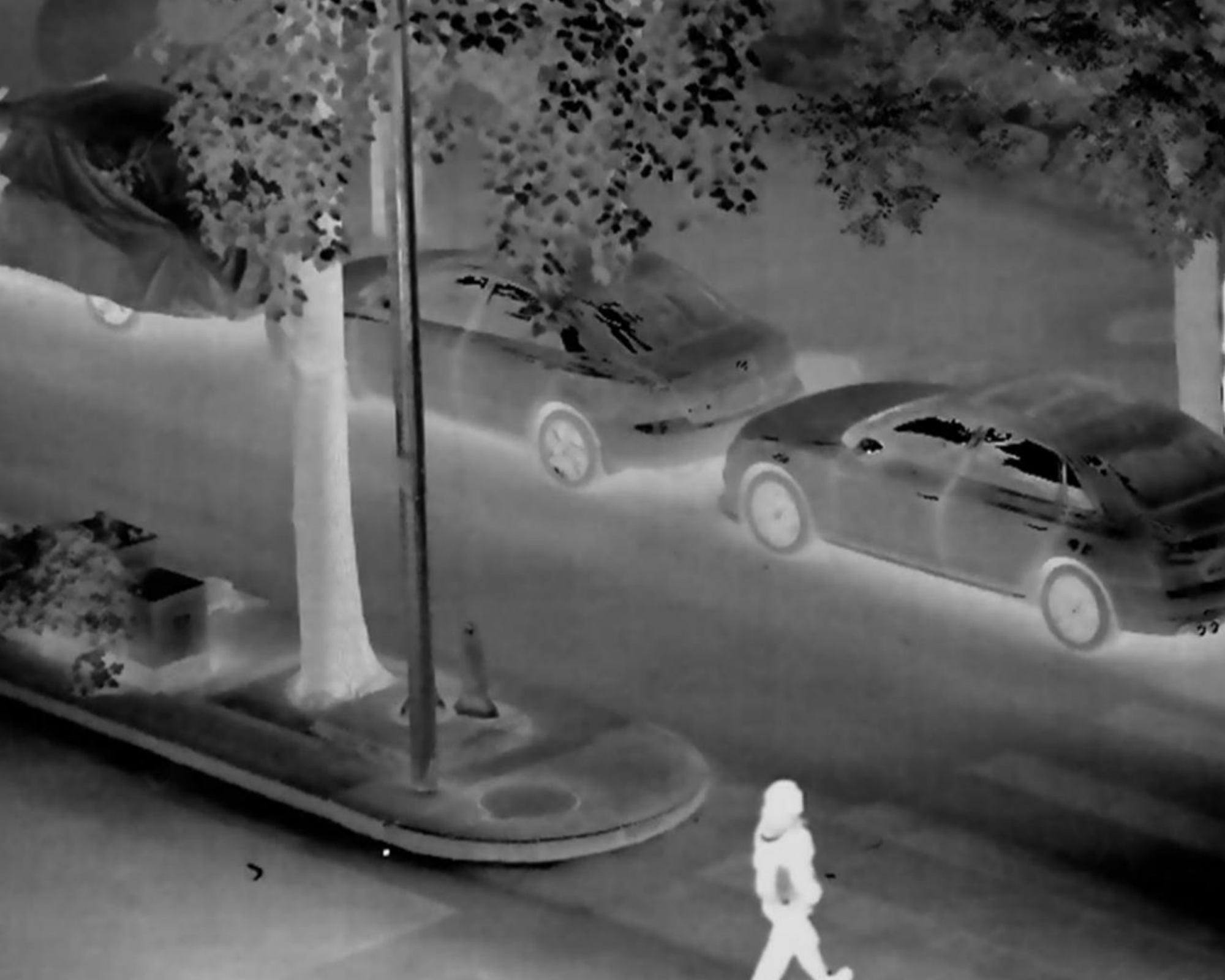}}
\end{tabular}
\begin{tabular}{c}
\begin{tikzpicture}
\draw[dashed, gray, line width=0.15pt] (0,0) -- (0,4.2);
\end{tikzpicture}
\end{tabular}
\begin{tabular}{c}
\subfloat[Saliency Map]{
\includegraphics[width=0.14\linewidth]{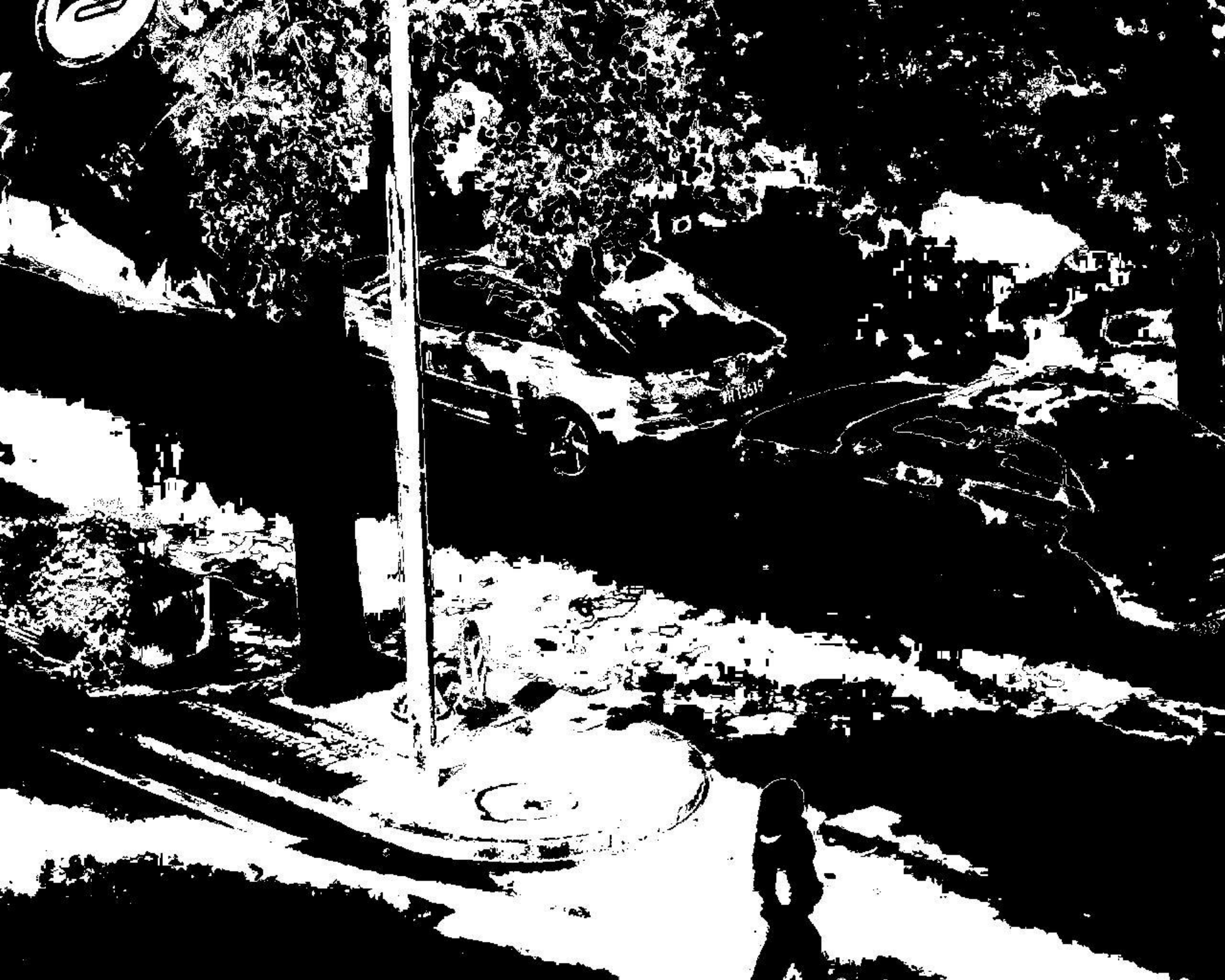}}
\subfloat[Fusion image]{
\includegraphics[width=0.14\linewidth]{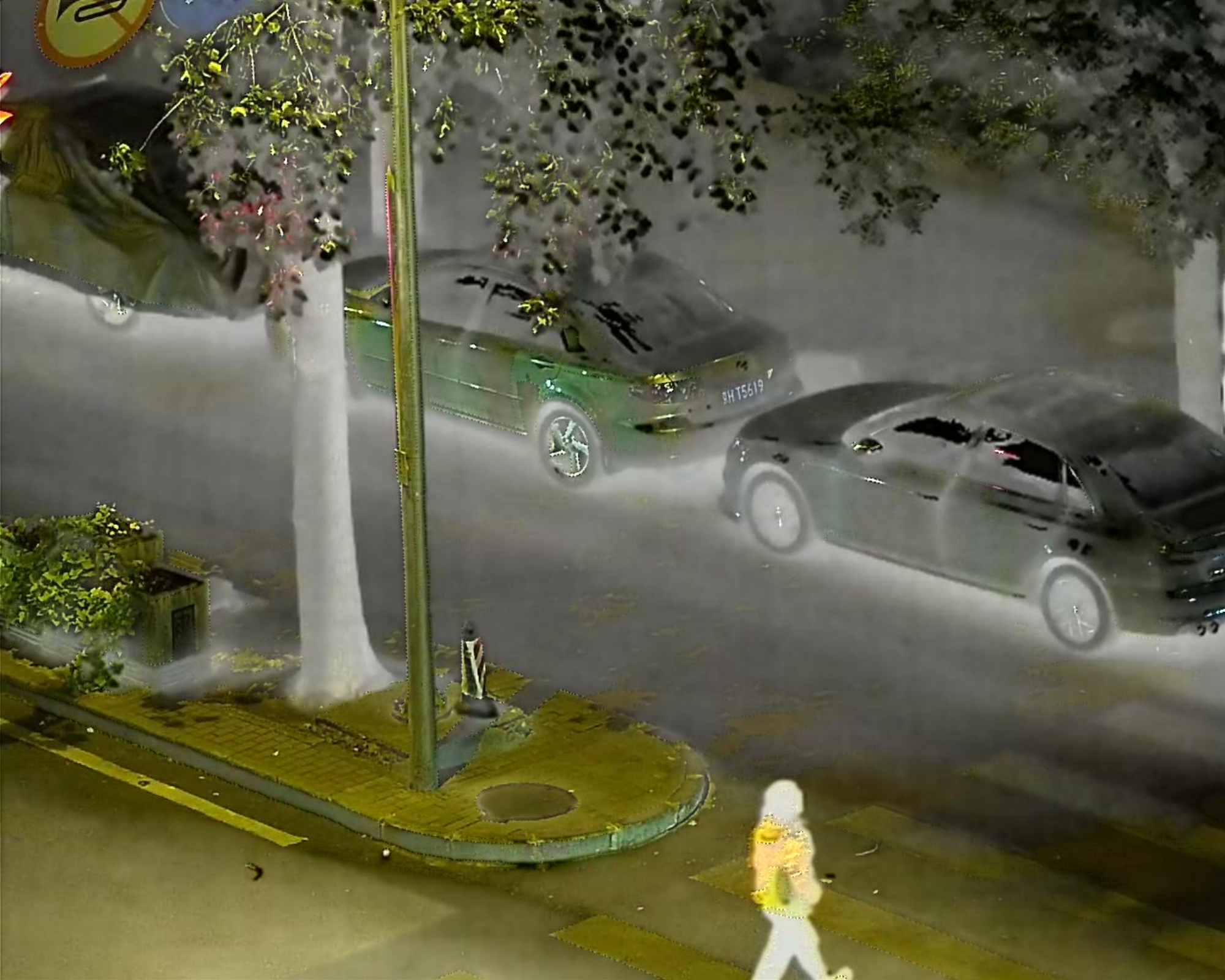}}
\\[2mm]
\subfloat[Y Channel]{
\includegraphics[width=0.14\linewidth]{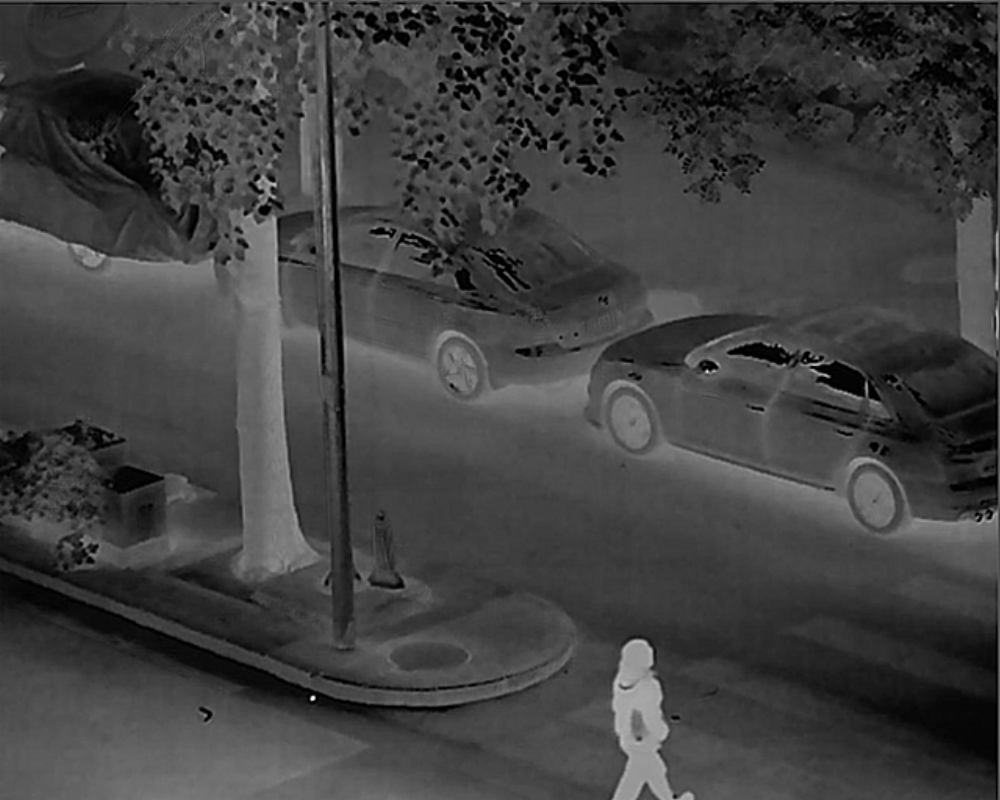}}
\subfloat[TDHF fusion]{
\includegraphics[width=0.14\linewidth]{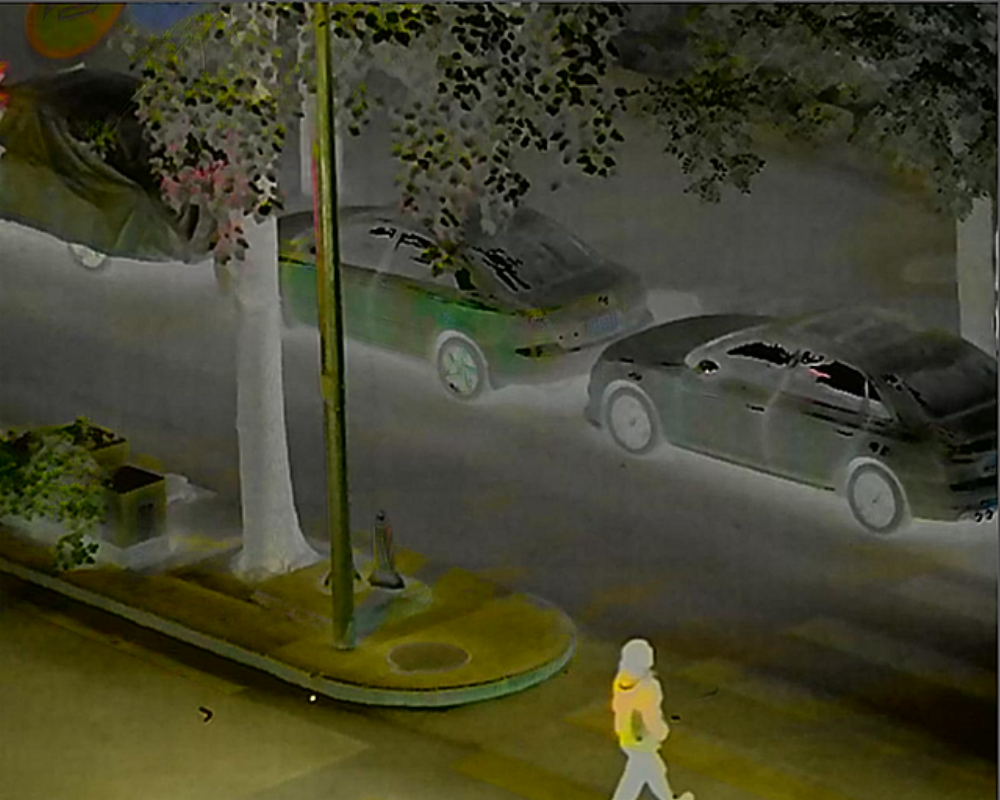}}
\end{tabular}
\begin{tabular}{c}
\begin{tikzpicture}
\draw[dashed, gray, line width=0.15pt] (0,0) -- (0,4.2);
\end{tikzpicture}
\end{tabular}
\begin{tabular}{ccc}
\subfloat[HF]{\includegraphics[width=0.14\linewidth]{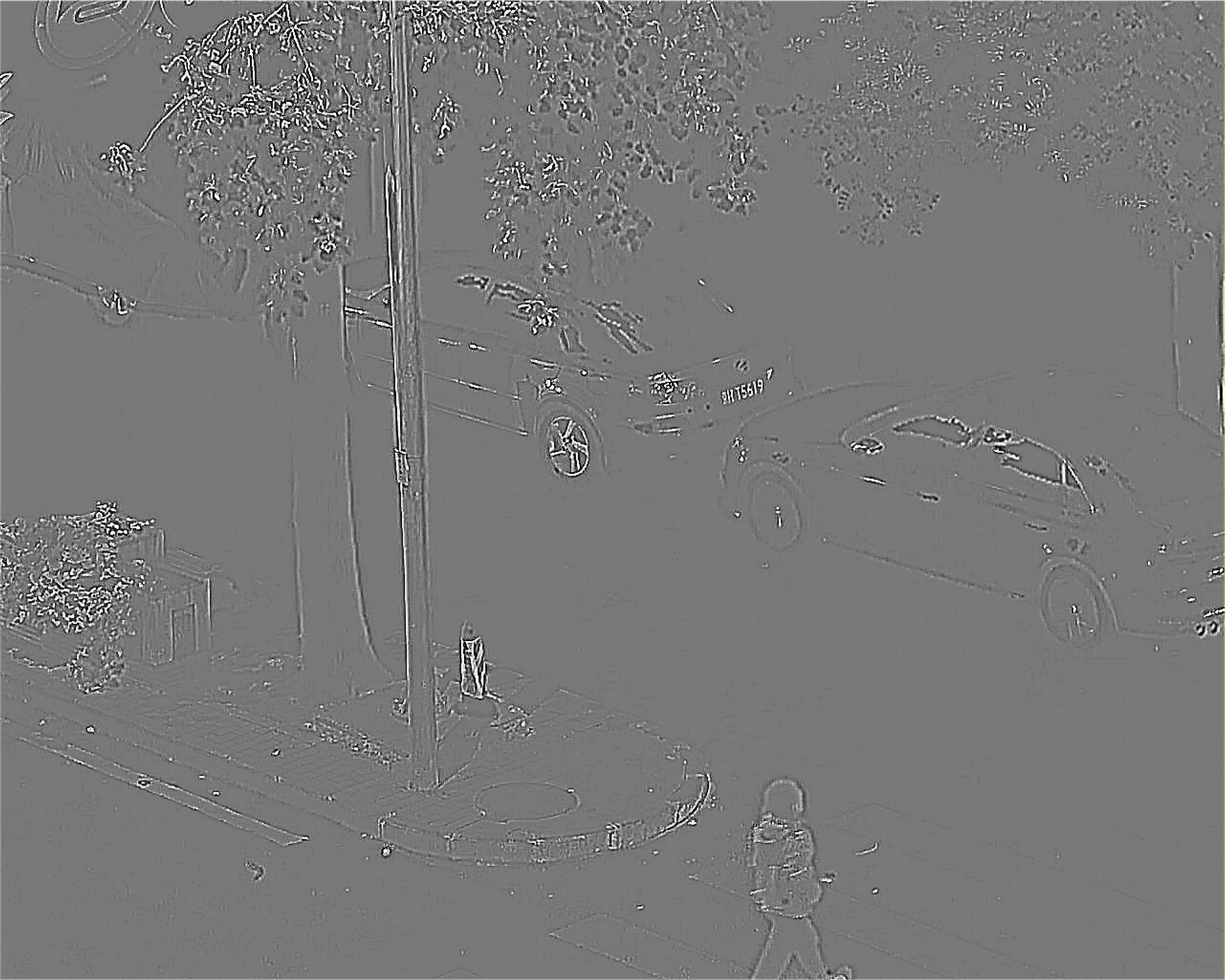}} &
\subfloat[LF]{\includegraphics[width=0.14\linewidth]{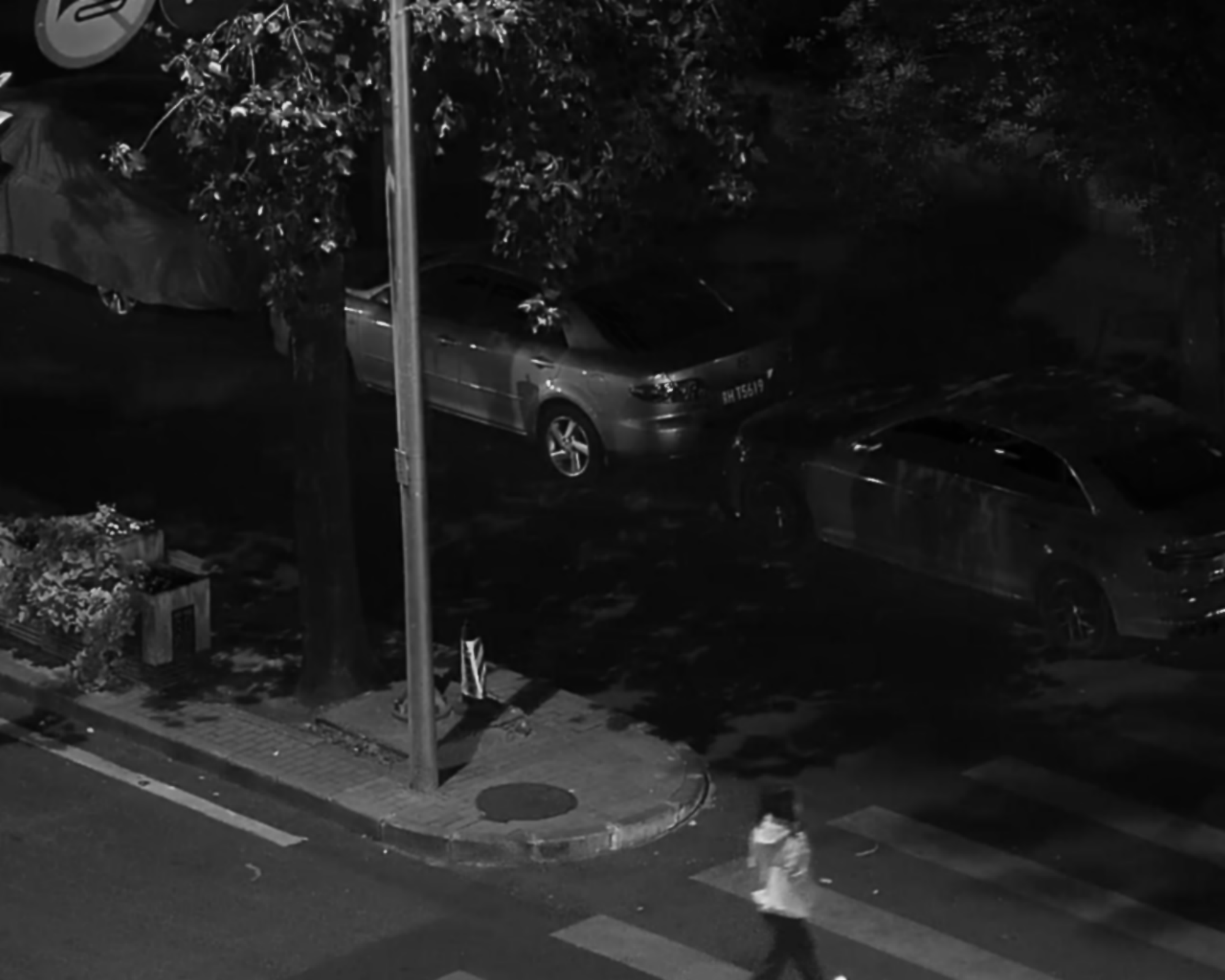}} &
\subfloat[MPC]{\includegraphics[width=0.14\linewidth]{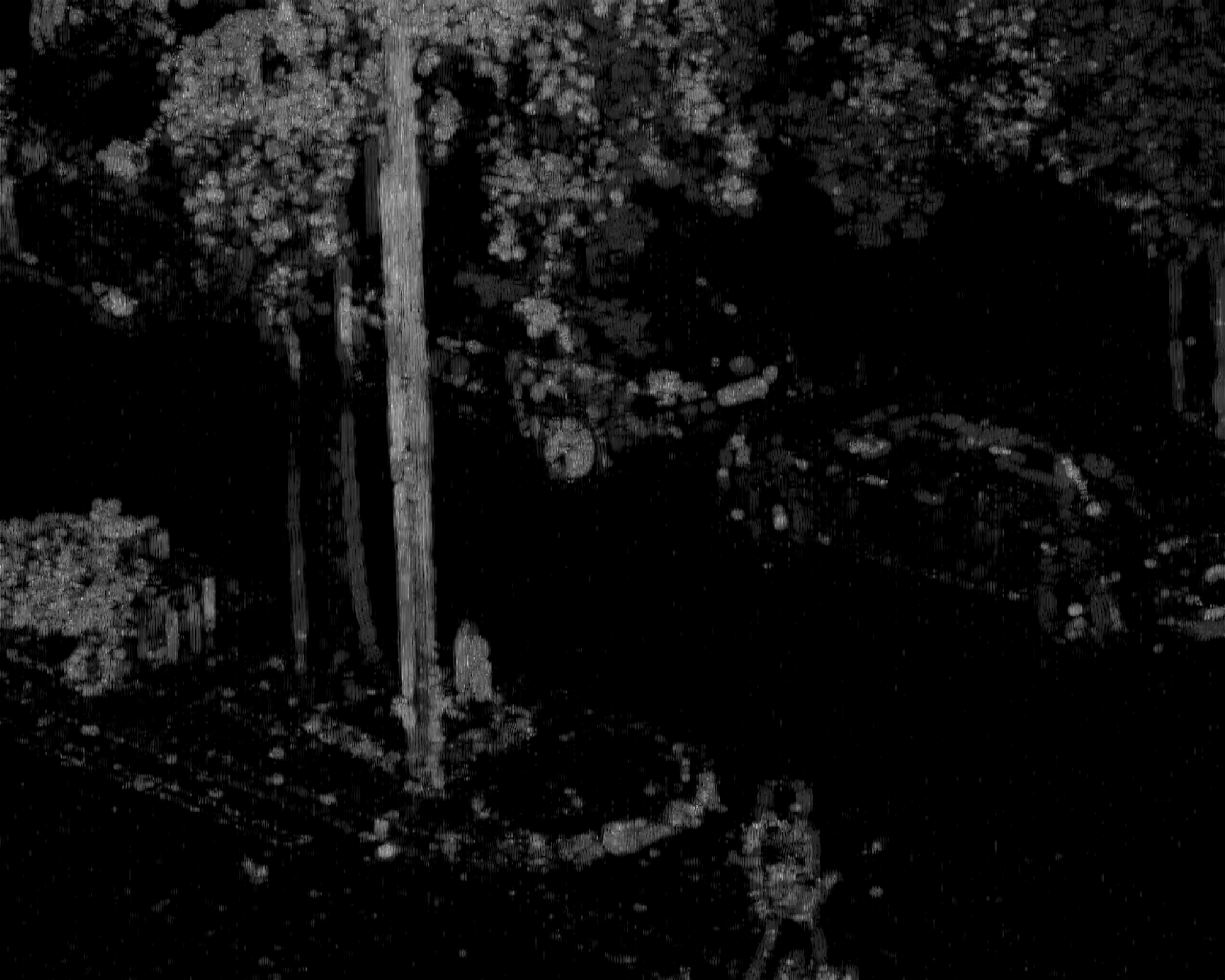}} \\[2mm]
\subfloat[T Map]{\includegraphics[width=0.14\linewidth]{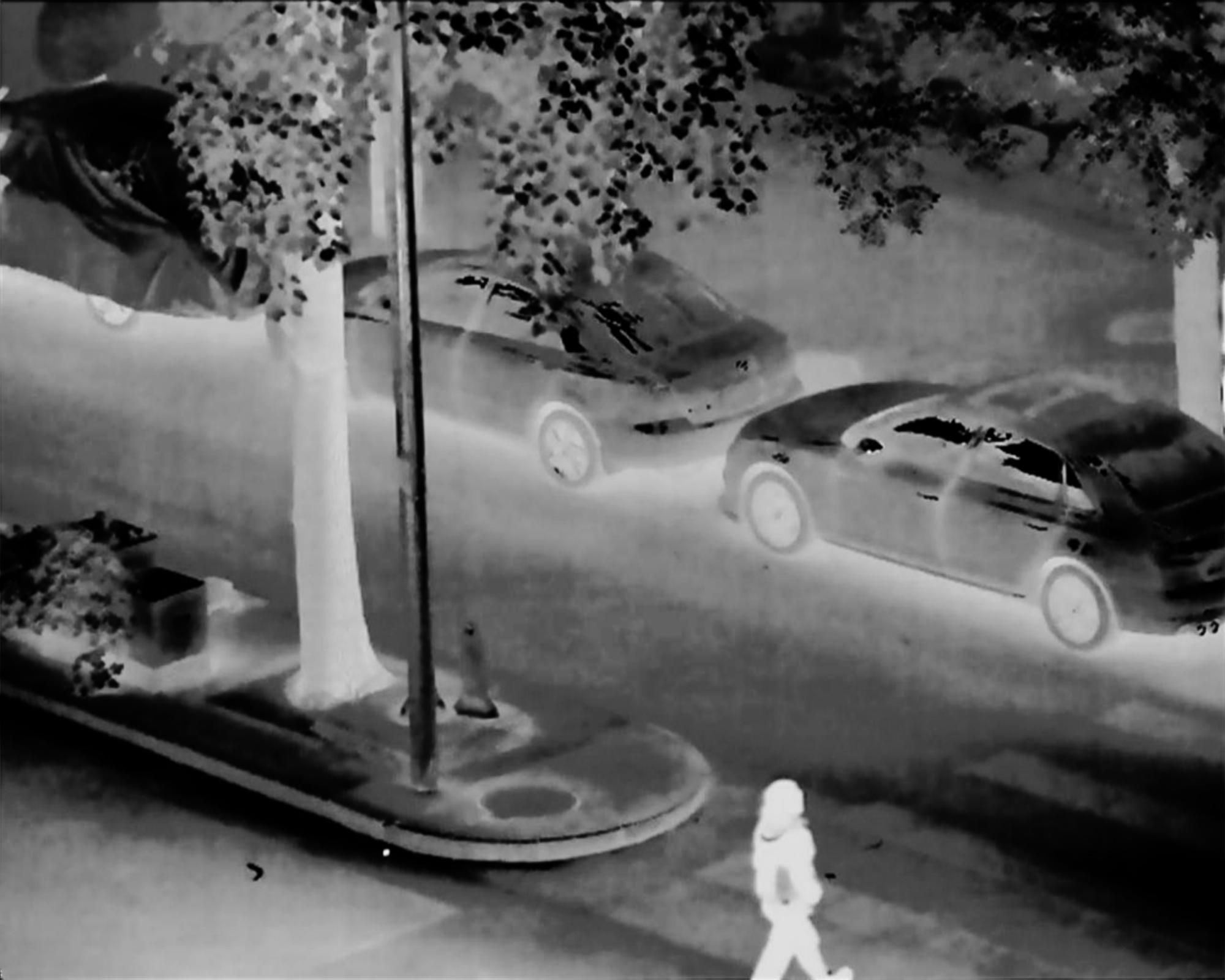}} &
\subfloat[e Map]{\includegraphics[width=0.14\linewidth]{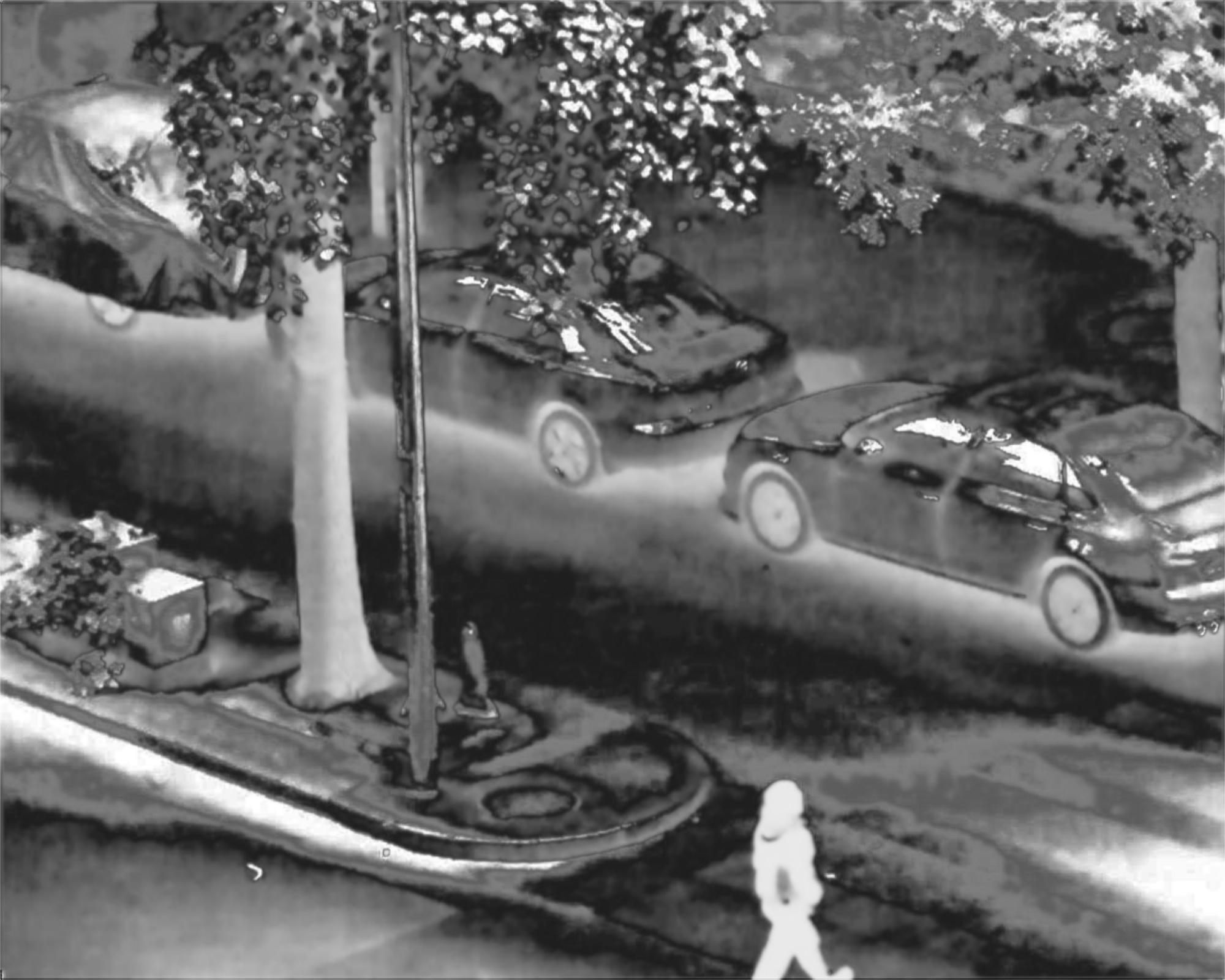}} &
\subfloat[X Map]{\includegraphics[width=0.14\linewidth]{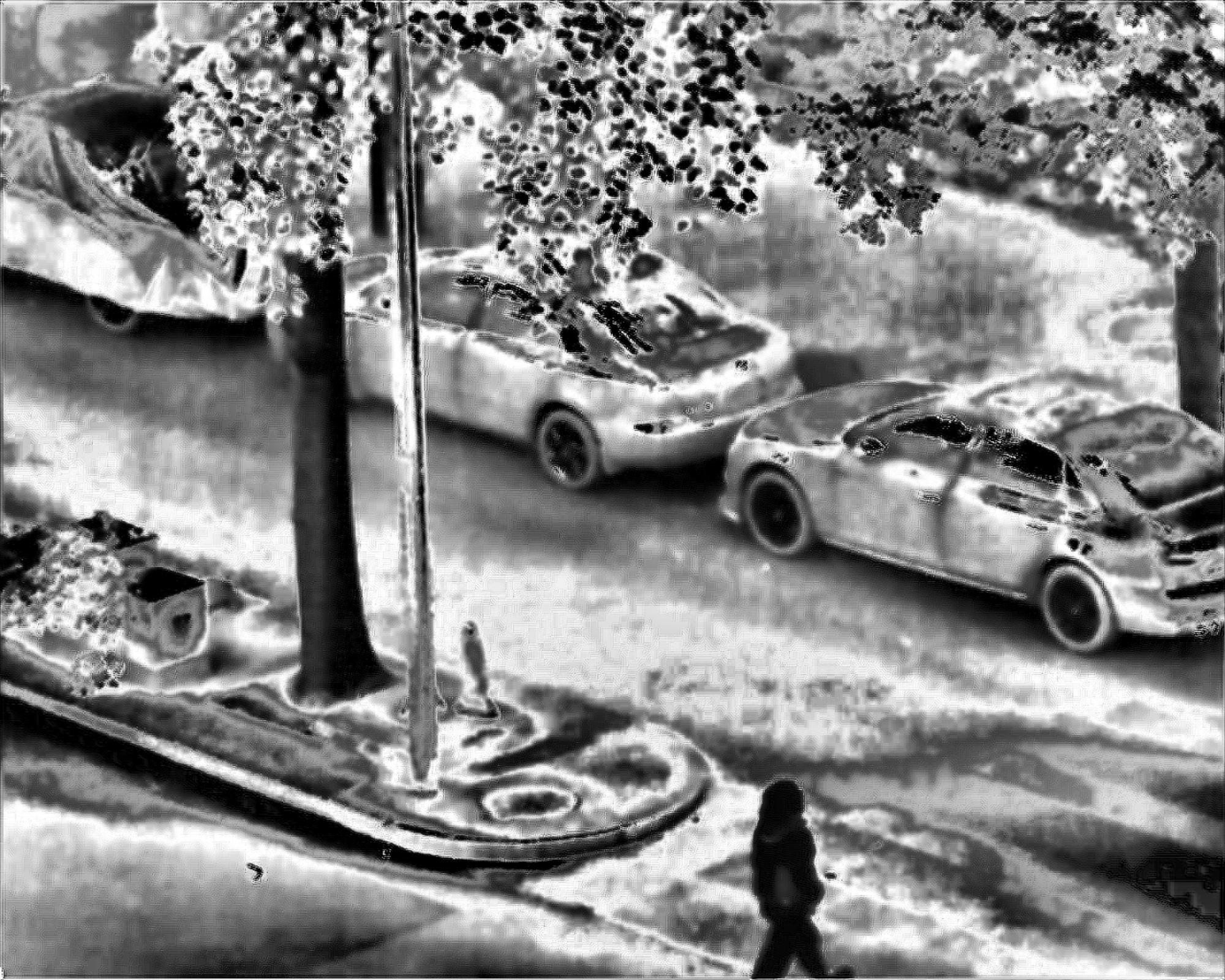}}
\end{tabular}
\end{tabular}
\caption{
The figure shows visible images, infrared images, saliency-based fusion, task-driven fusion, high-frequency maps, low-frequency maps, monogenic phase congruency maps, and T-Map, e-Map, and X-Map results under low-light conditions.
}
\label{Pic_Visualization_LLVIP}
\end{figure*}

\subsection{Ablation Study}

\begin{table}[t]
\centering
\caption{
Ablation study of the proposed MVR-MoE framework. 
Each component, including Mixture-of-Experts (MoE), Multi-View Structure (MS), Multi-View Hyper Connections (MVHC), and Expert Specialization (ES), is progressively introduced to evaluate its contribution. 
$\Delta$ denotes the improvement over the baseline.
}
\label{tab_ablation}
\setlength{\tabcolsep}{5pt}
\renewcommand{\arraystretch}{0.88}
\begin{tabular}{lcccc cr}
\toprule
Setting & MoE & MS & MVHC & ES & mAP@0.5:0.95 & $\Delta$ \\
\midrule
Baseline 
    & \xmark & \xmark & \xmark & \xmark  & 44.4 &  \\
+MoE      
    & \cmark & \xmark & \xmark & \xmark  & 45.4 & +1.0 \\
+MS      
    & \cmark & \cmark & \xmark & \xmark  & 46.8 & +1.4 \\ 
\hline
+MVHC      
    & \cmark & \cmark & \cmark & \xmark  & 48.0 & +1.2 \\
+ES      
    & \cmark & \cmark & \xmark & \cmark  & 47.7 & +0.9 \\
All     
    & \cmark & \cmark & \cmark & \cmark  & \textbf{51.8} & \textbf{+5.0} \\
\bottomrule
\end{tabular}
\end{table}

We conduct ablation studies to evaluate each component of MVR-MoE, as shown in Table~\ref{tab_ablation}.
The baseline directly predicts the $T$, $e$, and $X$ channels from fused features using three parallel branches, achieving 44.4\% mAP.
Introducing the mixture-of-experts (MoE) improves performance to 45.4\% by decomposing channel generation into multiple specialized experts that capture task-relevant patterns from shared features.
We further incorporate the multi-structure (MS) representation, which provides complementary cues such as high-frequency, low-frequency, and MPC-based phase information. This enriches the feature space and boosts performance to 46.8\% mAP, serving as the intermediate configuration for subsequent analysis.
Building on this, the multi-view hyper connections (MVHC) mechanism enables cross-branch feature interaction, enhancing multi-view feature diversity and improving performance to 48.0\% mAP.
We also explore expert selection (ES), which dynamically routes features to appropriate experts based on multi-modal inputs, achieving 47.7\% mAP.
Finally, combining MVHC and ES forms the full MVR-MoE, significantly improving performance to 51.8\% mAP.
While MVHC and ES individually yield moderate gains, their combination leads to a substantially larger improvement, indicating that the two components are complementary in enhancing feature representation and expert utilization.

\subsection{Visualization Analysis}

\begin{figure}[!t]
\centering
\includegraphics[width=0.9\linewidth]{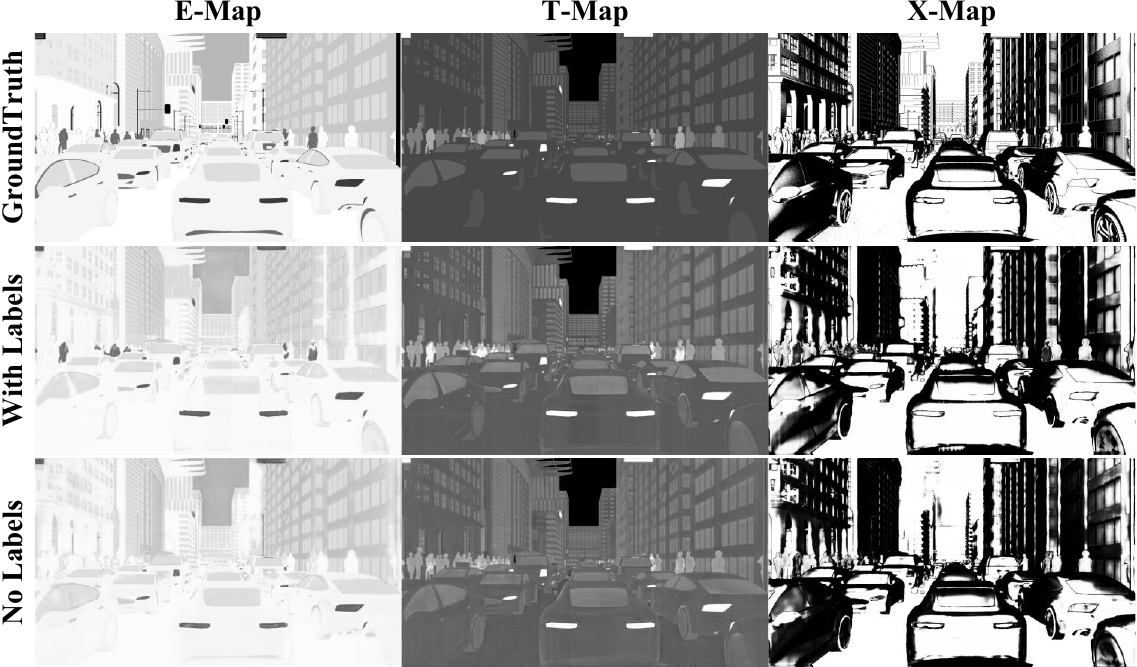}
\caption{
Ground truth and generated TeX maps under supervised and self-supervised settings on HADAR.
}
\label{Pic_TeV}
\end{figure}

We present the Multi-View Structural (HF, LF, and MPC), T-Map, e-Map, and X-Map results mined from infrared and visible images. 
As shown in Figure \ref{Pic_Visualization_LLVIP}, our method does not merely fuse infrared and visible inputs, but decomposes them into structurally meaningful multi-view representations. 
The high-frequency and low-frequency maps capture complementary local and global cues, whereas the monogenic phase congruency map highlights robust structural boundaries. 
Meanwhile, the T-Map, e-Map, and X-Map indicate that the model disentangles temperature, emissivity, and texture information in a physically meaningful manner. 
In contrast, saliency-based fusion and task-driven fusion primarily aggregate multimodal information into a single representation, which tends to blur fine-grained structures or bias toward dominant modalities. 
As a result, subtle details and material-dependent characteristics are not explicitly preserved. 
This design preserves fine details of humans, vehicles, and background objects under low-light conditions while enhancing discriminability and robustness across materials and scene components.
We additionally evaluate MVR-MoE on the simulated HADAR dataset, where ground-truth TeX annotations are available. 
As shown Figure~\ref{Pic_TeV}, the generated maps preserve the main physical structures and remain reasonably close to the corresponding target modalities, even without explicit label supervision. 

\subsection{Computational Complexity Analysis}

We further evaluate the computational efficiency of the proposed MVR-MoE in terms of inference time, FLOPs, and the number of parameters. 
For this evaluation, we randomly sample 10 images from the M3FD dataset, each with a resolution of $1024\times768$, and measure the average inference time per image on an NVIDIA GeForce RTX 4090 GPU. 
All results are averaged over multiple runs. 
As reported in Table~\ref{tab_speed_comparison}, the proposed method processes a single image in 29.03\,ms, with 374.7\,G FLOPs and only 0.487\,M parameters, MVR-MoE maintains a favorable balance between computational efficiency and representation capability. 
In particular, the proposed architecture introduces only a very small parameter footprint while enhancing feature representation through structural design, making it practical for efficient real-world deployment.

\begin{table}[t]
\centering
\caption{Computational Efficiency Comparison of Different Methods ($\downarrow$ indicates lower is better).}
\label{tab_speed_comparison}
\renewcommand{\arraystretch}{0.88}
\begin{tabularx}{\columnwidth}{X r r r}
\toprule
Method & Time$_{(ms)}$$\downarrow$ & FLOPS$_{(G)}$$\downarrow$ & Params$_{(M)}$$\downarrow$ \\
\midrule
TGFuse \cite{rao2023tgfuse}
    & 41.17  & 318.9  & 137.30 
\\
CDDFuse \cite{zhao2023cddfuse}
    & 463.8  & 1402.0  & 1.186 
\\
DDFM \cite{zhao2023ddfm}
    & 280.8k  & 1336k  & 552.700 
\\
IGNet \cite{li2023learning}
    & 47.46  & 647.6  & 7.871 
\\
MetaFusion \cite{zhao2023metafusion}
    & 85.40  & 637.9  & 0.812 
\\
MoE-Fusion \cite{cao2023multi}
    & 77.79  & \textbf{229.5}  & 83.450 
\\
PSFusion \cite{tang2023rethinking}
    & 79.66  & 463.6  & 45.900 
\\
SegMiF \cite{liu2023multi}
    & 681.50  & 500.4  & 0.621 
\\
MRFS \cite{zhang2024mrfs}
    & 98.90  & 356.0  & 135.000
\\
CoCoNet \cite{liu2024coconet}
    & 29.06  & 498.5  & 9.114 
\\
FILM \cite{zhao2024image}
    & 183.3  & 230.4  & 0.490 
\\
TDHF \cite{li2025tdhf}
    & 188.7  & 1135.71  & 57.1
\\ \hline
\textbf{MVR-MoE}
    & \textbf{29.03}  & 374.7  & \textbf{0.487} 
\\
\bottomrule
\end{tabularx}
\end{table}
\section{Conclusion}
In this work, we propose MVR-MoE, a physically consistent framework for TeX generation under low-light conditions. 
The framework organizes infrared and visible observations into multi-view structural representations, capturing both shared structural evidence and modality-specific characteristics. 
Multi-View Hyper Connections expand the structural feature space and enable cross-view interaction, while a hierarchical routing strategy adaptively regulates feature utilization at both the structural-view and expert levels. 
With Stefan-Boltzmann-law constraints, the predicted temperature, emissivity, and texture components remain consistent with their underlying radiative relationships, reducing artifacts from purely data-driven fusion. 
Extensive experiments demonstrate that MVR-MoE achieves strong discriminability and robustness with competitive efficiency. 
These results highlight the effectiveness of jointly modeling structural diversity and physically consistent for multimodal representation learning. 
The proposed framework connects physics-based modeling with multi-view representation learning and provides a reliable solution for TeX generation in complex environments. 
Nevertheless, sensitivity to emissivity estimation remains under complex conditions, future work will investigate more specialized emissivity modeling strategies to improve robustness and physical fidelity.

\section{Acknowledgments}

This work was funded by the Basic Research Program of Jiangsu under Grant No. BK20253028 and the National Natural Science Foundation of China under Grant Nos. U24A20330 and 62361166670.

\bibliographystyle{ACM-Reference-Format}
\balance
\bibliography{sample-base}

\end{document}